\documentclass[sigconf,screen]{acmart}
\usepackage{makecell}  
\usepackage{multirow}  
\usepackage{graphicx}
\usepackage{subfigure}
\usepackage{graphicx}
\usepackage{booktabs}
\usepackage{caption}
\usepackage{amsmath}
\usepackage{enumitem}
\usepackage{subcaption}
\AtBeginDocument{%
  }

\setcopyright{acmlicensed}
\copyrightyear{2018}
\acmYear{2025}
\acmDOI{XXXXXXX.XXXXXXX}
\acmConference[MM'25]{ACM Conference}{October 27--31,
  2025}{Dublin, Ireland}
\acmISBN{978-1-4503-XXXX-X/2018/06}

\acmSubmissionID{5585}
\renewcommand\footnotetextcopyrightpermission[1]{}
\begin{document}

\title{SAGE: A Visual Language Model for Anomaly Detection via Fact Enhancement and Entropy-aware Alignment}

\author{Guoxin Zang}
\affiliation{%
  \institution{Harbin Institute of Technology}
  \city{Harbin}
  \country{China}}
\email{gxzang@stu.hit.edu.cn}

\author{Xue Li}
\affiliation{%
  \institution{Harbin Institute of Technology}
  \city{Harbin}
  \country{China}}
\email{lixuecs@hit.edu.cn}

\author{Donglin Di}
\affiliation{%
  \institution{DZ Matrix}
  \city{Beijing}
  \country{China}}
\email{donglin.ddl@gmail.com}

\author{Lanshun Nie}
\affiliation{%
  \institution{Harbin Institute of Technology}
  \city{Harbin}
  \country{China}}
\email{nls@hit.edu.cn}

\author{Dechen Zhan}
\affiliation{%
  \institution{Harbin Institute of Technology}
  \city{Harbin}
  \country{China}}
\email{dechen@hit.edu.cn}

\author{Yang Song}
\affiliation{%
  \institution{University of New South Wales}
  \city{Sydney}
  \country{Australia}
}
\email{yang.song1@unsw.edu.au}

\author{Lei Fan}
\authornote{Corresponding author.}
\affiliation{%
  \institution{University of New South Wales}
  \institution{DZ Matrix}
  \city{Sydney}
  \country{Australia}
}
\email{lei.fan1@unsw.edu.au}

\renewcommand{\shortauthors}{Guoxin Zang, et al.}
\newcommand{\blue}[1]{\textcolor{blue}{#1}}
\begin{abstract}
While Vision-Language Models (VLMs) have shown promising progress in general multimodal tasks, they often struggle with industrial anomaly detection and reasoning, particularly in delivering interpretable explanations and generalizing to unseen categories. This limitation stems from the inherently domain-specific nature of anomaly detection, which hinders the applicability of existing VLMs in industrial scenarios that require precise, structured, and context-aware analysis. To address these challenges, we propose \textbf{SAGE}, a VLM-based framework that enhances anomaly reasoning through Self-Guided Fact Enhancement (SFE) and Entropy-aware Direct Preference Optimization (E-DPO). SFE integrates domain-specific knowledge into visual reasoning via fact extraction and fusion, while E-DPO aligns model outputs with expert preferences using entropy-aware optimization. Additionally, we introduce \textbf{AD-PL}, a preference-optimized dataset tailored for industrial anomaly reasoning, consisting of 28,415 question-answering instances with expert-ranked responses. To evaluate anomaly reasoning models, we develop Multiscale Logical Evaluation (\textbf{MLE}), a quantitative framework analyzing model logic and consistency. SAGE demonstrates superior performance on industrial anomaly datasets under zero-shot and one-shot settings. The code, model, and dataset are available at \url{https://github.com/amoreZgx1n/SAGE}.
\end{abstract}




\maketitle

\begin{figure}[t]
  \centering
  \includegraphics[width=\linewidth]{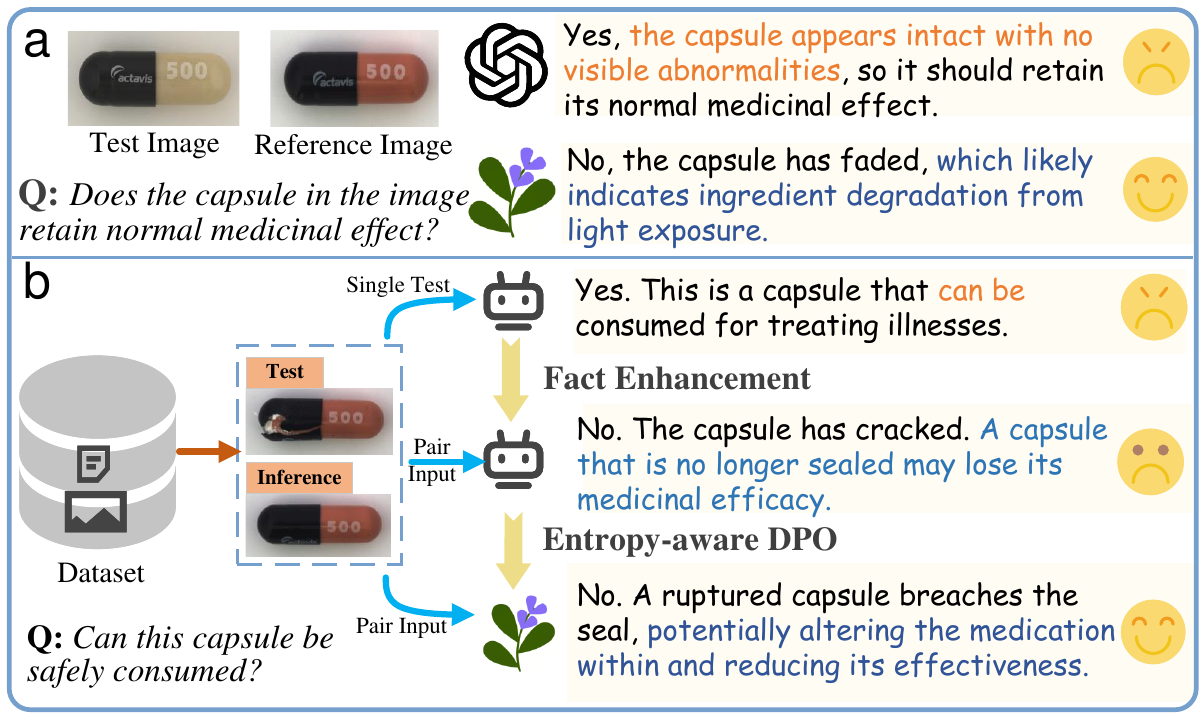}
  \caption{a. Comparison between our SAGE and GPT-4o. SAGE identifies the uncommon anomaly and provides the correct answer. b. An illustration of SAGE: a VLM using fact enhancement and preference optimization with comparison learning.}
  \label{label:begin figure}
  \Description{The begin figure}
\end{figure}

\section{Introduction}
Industrial visual anomaly detection plays a crucial role in ensuring product quality, operational safety, and cost efficiency in manufacturing and maintenance systems \cite{li2024towards, wang2023multimodal, yang2025dsdnet, yang2025learning}. Early approaches \cite{defard2021padim, reiss2021panda} typically formulate anomaly detection as an unsupervised classification or segmentation problem, aiming to determine whether an anomaly exists and to localize the anomalous regions. Existing methods can be broadly categorized into three types: Embedding-based methods (\textit{e.g.}, SVDD \cite{tax2004support}, PatchCore \cite{roth2022towards}) extract feature embeddings using pre-trained models \cite{he2016deep, dosovitskiy2020image,su2021k} and learn boundaries for normal samples to detect deviations; Data augmentation-based methods (\textit{e.g.}, CutPaste \cite{li2021cutpaste}, DRAEM \cite{ezeme2019dream}) synthesize pseudo-anomalies through various data transformation strategies \cite{cimpoi2014describing, perez2023poisson}, enabling supervised learning for anomaly detection; Reconstruction-based methods (\textit{e.g.}, MemAE \cite{gong2019memorizing}, RD \cite{deng2022anomaly}) adopt a compression–reconstruction paradigm, learning the patterns of normal samples and identifying anomalies based on reconstruction error.

More recently, with the advancement of vision-language models, methods such as AnomalyCLIP \cite{zhou2023anomalyclip} and WinCLIP \cite{jeong2023winclip} leverage CLIP's \cite{radford2021learning} zero-shot or few-shot capabilities to extract aligned embeddings for images and text descriptions, exploring the use of image–text pairing with anomalous keywords. Beyond CLIP, Vision-Language Models (VLMs) \cite{touvron2023llama, zheng2023judging}, with larger architectures and enhanced multimodal reasoning capabilities, have also been investigated for anomaly detection \cite{gao2024vision, cai2024anomaly, yang2024follow, xu2025towards,jiang2024mmad}. Although these methods have achieved remarkable progress in identifying known anomalies within curated categories, they often rely heavily on large quantities of normal samples for each specific class. As a result, they typically lack the ability to provide interpretable reasoning and to generalize to unseen categories in long-tail industrial environments. 

In real-world industrial scenarios, anomaly detection often requires deeper reasoning capabilities to answer what (\textit{e.g.}, anomaly types), why (\textit{e.g.}, root causes), and how (\textit{e.g.}, severity and downstream implications). For instance, detecting a defective component on an assembly line may not only involve localization but also diagnosing the potential cause, prioritizing corrective actions, and anticipating its impacts on subsequent processes. Advanced VLMs offer a preliminary form of anomaly reasoning through their interactive dialogue capabilities \cite{gu2023anomalyagpt}. However, their performance is often constrained by the diverse and large-scale web-based pretraining data, which may conflict with the domain-specific definitions of normality required in industrial contexts. As illustrated in Figure~\ref{label:begin figure}(a), a capsule is incorrectly classified as normal by GPT-4o, as the model relies on generic commonsense patterns (\textit{e.g.}, detecting cracks or spots), while failing to recognize domain-specific anomalies (\textit{e.g.}, color inconsistencies) that are critical in industrial quality control. In contrast, human experts can quickly learn the characteristics of normal samples by leveraging prior experience and commonsense reasoning. These limitations highlight the need for advanced frameworks that effectively integrate visual evidence with domain-specific knowledge, enabling interpretable, generalizable, and actionable anomaly detection frameworks.

To address these limitations, we aim to identify anomalies through \textbf{Comparison-based Learning}, where the model analyzes the test image in relation to a given normal reference image that provides patterns of normality and contextual cues. Most existing VLMs \cite{touvron2023llama, zheng2023judging}, pre-trained primarily on single-image inputs, struggle to accurately interpret multi-image comparisons. To mitigate this, we design specific prompts that first guide the model to attend to the reference image in order to establish an expected baseline, thereby enabling more accurate reasoning about potential semantic deviations in the test image. Considering that VLMs lack the domain experience of human experts, we further incorporate preference learning \cite{rafailov2023direct} and fact enhancement \cite{zeng2025siftgroundingllmreasoning}, which leverage expert rankings of different reasoning outcomes to align the model’s responses with domain-specific criteria. As a result, the model can more effectively interpret reference images and generate higher-quality responses, aligned with expert expectations.

To this end, we propose a VLM-based framework for anomaly detection and reasoning, termed SAGE, which fine-tunes a pretrained VLM model with two novel strategies: Self-Guided Fact Enhancement (SFE) and Entropy-aware DPO (E-DPO) alignment. Specifically, unlike existing VLM-based methods that directly identify anomalies based on a single test image, our approach adopts a comparison-based input paradigm, where a test image is paired with a reference image sampled from normal classes. During the SFE, we first extract fact information (\textit{e.g.,} key characteristics and conditions) from the test image leveraging pretrained VLM, guided by a customized prompt template designed to summarize image content. These facts are then fused with the visual features of both input images via a cross-attention mechanism, which is subsequently fed into the LLM to perform anomaly reasoning.
By using E-DPO, the VLM is further tuned to align its outputs with expert preference distributions. This is achieved via entropy-aware direct preference optimization, which leverages the entropy differences across multiple ranked response candidates generated for each input image pair. To support this, we further construct a new multiple-choice dataset, \textbf{AD-PL}, comprising 28,415 preference-based QA instances. Each instance includes a test–reference image pair along with four candidate answers ranked by domain experts. To comprehensively evaluate the quality of anomaly reasoning responses, we propose Multiscale Logical Evaluation (MLE), which leverages GPT-4o to quantitatively measure performance across three key dimensions: identification accuracy, localization precision, and reasoning consistency.

Our key contributions can be summarized as follows: 
\begin{itemize}[topsep=0pt, parsep=0pt, itemsep=1pt]
\item We propose \textbf{SAGE}, a VLM-based framework for anomaly detection and reasoning. SAGE employs comparison-based learning by utilizing a normal image as a reference to enable deep anomaly reasoning on a given test image.

\item We design two fine-tuning strategies: SFE, which incorporates domain-specific knowledge for visual reasoning, and E-DPO, which aligns model responses with expert preferences.

\item We construct a high-quality dataset with 28,415 QA instances, each with four ranked candidate answers for industrial anomaly reasoning.

\item We develop a comprehensive quantitative evaluation framework for anomaly reasoning models. Our method exhibits superior performance under both zero-shot and one-shot evaluation settings on the MANTA \cite{fan2024manta} and MPDD \cite{9631567} datasets.
\end{itemize}

\begin{figure*}[t] 
    \centering
    \includegraphics[width=\textwidth]{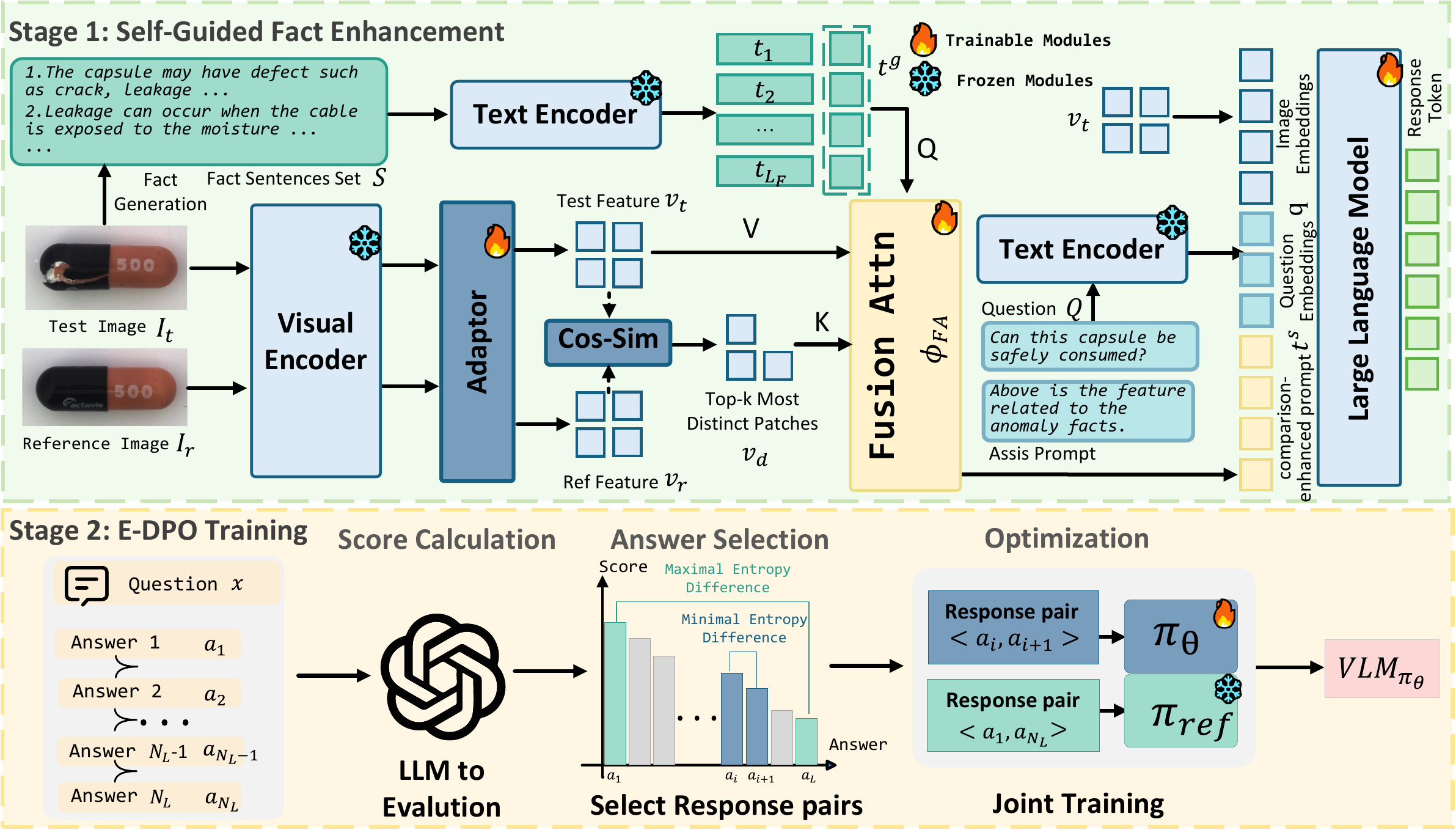} 
    \caption{
    The overview of SAGE. It consists of two training strategies: Self-Guided Fact Enhancement (SFE) and Entropy-aware Direct Preference Optimization (E-DPO). SFE fine-tunes the VLM (including the adapter, fusion attention layer, and LLM) based on comparison-based learning, in which the feature differences between the test and reference images are fused with extracted fact sentences using a cross-attention mechanism. E-DPO utilizes multiple preference-ranked answers to select response pairs with the largest entropy difference among candidates; during this stage, the LLM and adapter are fine-tuned accordingly.} 
    \label{fig:main} 
\end{figure*}

\section{Related Work}
\textbf{Anomaly Detection.} Conventional anomaly detection (AD) methods \cite{scholkopf1999support} typically learn dataset-specific normal patterns via paradigms like one-class classification \cite{cohen2020sub}, reconstruction \cite{deng2022anomaly, fan2025grainbrain,fan2024revitalizing}, data augmentations~\cite{fan2024patch, fan2023identifying}, feature embedding comparison \cite{bae2023pni,liang2025tocoad}, or knowledge distillation \cite{bergmann2020uninformed}. More recently, pre-trained Vision-Language Models (VLMs), especially CLIP \cite{radford2021learning}, have gained attention due to their generalization capabilities. These VLM-based approaches adopt various strategies, including comparing image embeddings against textual descriptions of ``normal'' and ``abnormal'' in the CLIP embedding space (\textit{e.g.}, WinCLIP \cite{jeong2023winclip}), often enhanced via attention mechanisms or window-based comparisons \cite{yang2023self, mishra2021vt}, adapting VLMs for AD tasks via fine-tuning \cite{wu2024vadclip} or prompt learning \cite{yao2024cpt}, and leveraging foundation models such as Grounding DINO \cite{zhang2022dino} and SAM \cite{kirillov2023segment} for text-prompted anomaly localization. Some methods such as AnomalyGPT \cite{gu2023anomalyagpt} further integrate Vicuna~\cite{vicuna2023} to facilitate reasoning, localization, and interpretability.

In contrast to the above methods, SAGE adopts a comparison-based paradigm by utilizing paired test–reference images, where the reference image represents normal patterns. This design not only improves anomaly localization accuracy but also provides explicit normal references, making it particularly well-suited for domain-specific industrial applications.

\noindent \textbf{Vision-Language Models.} Recent advancements in VLMs \cite{chen2024internvl, gupta2022towards, liu2023visual, li2022blip, alayrac2022flamingo, Qwen-VL, wu2024detecting} have significantly enhanced multimodal understanding by integrating powerful vision encoders with Large Language Models (LLMs) \cite{chowdhery2023palm, touvron2023llama}. Foundational models such as CLIP \cite{radford2021learning}, pre-trained on large-scale image–text pairs, exhibit strong zero-shot generalization capabilities and have influenced a wide range of downstream tasks. The development of Multi-modal Large Language Models (MLLMs), such as GPT-4o \cite{hurst2024gpt} and open-source alternatives \cite{wu2024deepseekvl2mixtureofexpertsvisionlanguagemodels}, further enhances reasoning and pushes the boundaries of visual–textual reasoning and interaction. These models have been applied to various tasks, including detection \cite{carion2020end}, video analysis \cite{arnab2021vivit}, and visual reasoning \cite{li2023blip, alayrac2022flamingo}. Their capabilities are continually improved through techniques such as contrastive prompt learning \cite{yao2024cpt} and instruction tuning \cite{liu2023visual}, supported by large-scale resources like LAION \cite{schuhmann2022laionb} and OpenCLIP \cite{ilharco_gabriel_2021_5143773}.

However, most existing VLMs are not inherently designed for anomaly detection and often lack the ability to perform comparison-based reasoning. In contrast, our method fine-tunes VLMs to anomaly domains by integrating domain-specific commonsense while also aligning model reasoning with expert preferences.

\section{Method}
\subsection{Overview}

The SAGE framework is designed to enhance the anomaly reasoning capabilities of the VLM, addressing the challenges of domain-specific anomaly detection and comparison-based learning. Built upon a pretrained VLM backbone (\textit{e.g.}, InternVL2~\cite{chen2024internvl}), SAGE consists of four main components: a visual encoder $\mathcal{F}_v(\cdot)$ with an Adapter for extracting image features, a text encoder $\mathcal{F}_t(\cdot)$ for processing textual inputs, a Fusion Attention layer $\phi_{FA}$ for capturing comparison embeddings between test and reference images, and an integrated LLM for multimodal reasoning and response generation, as illustrated in Figure~\ref{fig:main}. In contrast to conventional VLMs \cite{chen2024internvl, gupta2022towards, liu2023visual}, SAGE adopts a test-reference image paradigm, and we propose two fine-tuning strategies: Self-Guided Fact Enhancement (SFE) and Entropy-aware Direct Preference Optimization (E-DPO) to adapt pretrained VLMs for anomaly detection and reasoning.

Given a test-reference image pair $\langle I_t, I_r \rangle$ where both $I_t, I_r \in \mathbb{R}^{H \times W \times 3}$, along with an input question $\mathbf{Q}$ of sequence length $L_Q$, the goal is to fine-tune a VLM to generate a textual response that includes both an anomaly detection decision and an explanation.  Specifically, the visual encoder $\mathcal{F}_v(\cdot)$ adopts a Vision Transformer (ViT) \cite{dosovitskiy2020image,su2022vitas} to extract patch-level embeddings \( v_t, v_r\in \mathbb{R}^{p^2 \times C} \) from the test and reference images \( \langle I_t, I_r \rangle \), respectively, where $p^2$ denotes the number of patches and \( C \) is the embedding dimension. To enable comparison-based learning within the VLM, the input question $\mathbf{Q}$ is augmented with an auxiliary instructional prompt: ``\textit{Above is the feature related to the anomaly facts}''. The combined input is then encoded by the text encoder $\mathcal{F}_t(\cdot)$ into sequence embeddings $\mathbf{q} \in \mathbb{R}^{L_q \times C}$. Subsequently, we employ the original pretrained VLM to generate $N$ textual fact sentences that describe potential anomalies and image patterns for each test image. This process is conducted offline and can be completed before fine-tuning the VLM. 

During the SFE stage, for each test-reference image pair, we retrieve the corresponding generated fact sentences and encode them using the text encoder $\mathcal{F}_t(\cdot)$ to obtain the fact embeddings $\mathbf{t}^g \in \mathbb{R}^{L_F \times C}$. To enable comparison-based learning, the top-$K$ patches are selected from the test image embedding $v_t$ based on their minimum cosine similarity to the reference image features $v_r$, forming the comparison visual embeddings $\mathbf{v}_d \in \mathbb{R}^{K \times \frac{H}{p} \times \frac{W}{p} \times C}$. These embeddings are then used in the Fusion Attention layer $\phi_{FA}$ to generate a comparison-enhanced prompt $\mathbf{t}^s \in \mathbb{R}^{N \times C}$ via a cross-attention mechanism, where $\mathbf{t}^g$ serves as the query (Q), $\mathbf{v}_d$ as the key (K), and $\mathbf{v}_t$ as the value (V). This Fusion Attention mechanism is employed to enhance the VLM’s awareness of test-reference discrepancies. Finally, the test image embeddings $\mathbf{v}_t$, the question embeddings $\mathbf{q}$, and the comparison-enhanced prompt $\mathbf{t}^s$ are concatenated and fed into the LLM to generate the textual response.

Next, during the E-DPO stage, the VLM is further refined to enhance its reasoning capabilities by leveraging multiple preference-ranked answers. The model is trained to align with expert preferences through entropy-aware optimization. E-DPO includes three steps: Score Calculation using an LLM, Answer Selection based on entropy differences, and model optimization. Given a test-reference image pair $\langle I_t, I_r \rangle$ and the corresponding question $\mathbf{Q}$, we provide a set of $L_{N}$ rank answers $(a_1,\cdots, a_{L_N})$ annotated by domain experts. We first perform the Score Calculation by leveraging an LLM (\textit{e.g.,} GPT-4o) to compute Multiscale Logical Evaluation (MLE) scores for each ranked answer, serving as a proxy for answer quality. Then, the Answer Selection selects informative answer pairs based on their entropy differences. Finally, these selected pairs are used to fine-tune the policy model $\pi_\theta$ (\textit{i.e.}, the Adapter in the visual encoder and LLM) through contrastive preference alignment.

\subsection{Self-Guided Fact Enhancement}

Although existing VLMs are capable of offering insightful reasoning for common anomalies, they often struggle to generalize to unseen categories or rare semantic anomalies. To address this limitation, we introduce the Self-Guided Fact Enhancement (SFE) module, which enables the VLM to better adapt to the anomaly detection domain. 

Given a test-reference image pair $\langle I_t, I_r \rangle$, the pretrained VLM is used to generate a set of $L_F$ fact sentences $s = \{s_1, s_2, \ldots, s_{L_F}\}$ that describe object-related characteristics, such as object categories, potential anomaly types, and relevant semantic facts. These sentences are produced using our predefined prompts, for example, ``\textit{What kinds of anomalies does this type of object tend to exhibit?}'' Each fact sentence $s_i$ is then encoded by the text encoder $\mathcal{F}_t(\cdot)$ into a fact embedding $\mathbf{t}_i$, forming the fact embedding set $\mathbf{t} = \{\mathbf{t}_1, \mathbf{t}_2, \ldots, \mathbf{t}_{L_F}\}$. A global pooling operation is subsequently applied to each sentence embedding to obtain its corresponding pooled representation $\mathbf{t}^g$.

For a given test-reference image pair $\langle I_t, I_r \rangle$, the visual encoder $\mathcal{F}_v(\cdot)$ is used to extract patch-level embeddings $\mathbf{v}_t$ and $\mathbf{v}_r$ from the test and reference images, respectively. To identify the most anomalous regions, we compute the cosine similarity between each patch in $\mathbf{v}_t$ and all patches in $\mathbf{v}_r$, and select the $K$ patches in the test image that exhibit the minimum similarity to any reference patch. These selected features, denoted as $\mathbf{v}_d \in \mathbb{R}^{K \times C}$, are defined as:
\begin{equation}
  \mathbf{v}^d = \underset{v_i \in v_t}{\text{argtop-$K$}} \left( \min_{v_j \in \mathbf{v}'} \text{sim}(v_i, v_j) \right),
  \label{eq:important}
\end{equation}
where $\text{sim}(v_i, v_j)$ denotes the cosine similarity between patch features $v_i$ and $v_j$, and the $\text{argtop-}K$ operator selects the $K$ patches in $\mathbf{v}_t$ with the lowest similarity scores. This operation highlights regions in the test image that are most dissimilar to the reference, and thus more likely to exhibit anomalous patterns.

Next, the Fusion Attention layer $\phi_{\text{FA}}$ is applied to obtain the comparison-enhanced prompt $\mathbf{t}^s$ by integrating the test image embeddings $\mathbf{v}_t$ and the selected patch-level differences $\mathbf{v}_d$, conditioned on the fact embeddings $\mathbf{t}^g$. This attention mechanism is formulated as:
\begin{equation}
    \mathbf{t}^s=
  \phi_{\text{FA}}(t^g, v_t, v_d) = \text{softmax}\left( \frac{t^g W_q (v_d W_k)^T}{\sqrt{d}} \right) v_t W_v,
  \label{eq:important}
\end{equation}
where $W_q$, $W_k$, and $W_v$ denote the learnable projection matrices for queries, keys, and values, respectively.  A multi-head attention mechanism is employed to enhance the interaction between fact sentences and visual patch embeddings. The attention operation is performed between each fact sentence and the corresponding comparison patches, and the outputs are stacked to yield the final comparison-enhanced embeddings $\mathbf{t}^s$.

Finally, the comparison-enhanced embeddings $\mathbf{t}^s$, the test image embeddings $\mathbf{v}_t$, and the question embeddings $\mathbf{t}^q$ are concatenated and fed into the LLM to generate the final answer. By leveraging fact priors and visual discrepancies, SFE effectively adapts the pretrained VLM to the anomaly detection domain, enhancing its interpretability and generalization to unseen cases.

\subsection{Entropy-aware DPO}

To address the limited anomaly-domain expertise in current VLMs and enhance their deeper reasoning capabilities in complex anomaly scenarios, we introduce an entropy-aware Direct Preference Optimization (E-DPO) strategy based on preference learning \cite{rafailov2023direct}. Considering the fact that human experts are able to recognize both normal and anomalous patterns while explaining root causes and downstream effects by drawing on prior experience and domain knowledge, and inspired by the instrumental fine-tuning approaches~\cite{liu2023visual}, we hypothesize that anomaly reasoning preference learning can be shared across diverse anomaly fields and encoded through unified learning. To support this optimization objective, we construct a preference-ranked anomaly reasoning dataset, AD-PL (described in Sec.~\ref{sec:dataset}), which serves as supervision for fine-tuning the model toward expert-aligned reasoning.

E-DPO consists of three steps: Score Calculation, Answer Selection, and Optimization. In the first step, Score Calculation, we leverage a large language model $\Gamma_{\text{LLM}}$ (e.g., GPT-4o) to evaluate each candidate answer $a_i$ along three dimensions: defect type identification, defect localization, and logical reasoning for a given question $x$. These aspects are used to compute a unified Multiscale Logical Evaluation (MLE) score:
\begin{equation}
  S_i = \frac{1}{3} \left( \Gamma_{\text{LLM}}^{\text{type}}(a_i) + \Gamma_{\text{LLM}}^{\text{loc}}(a_i) + \Gamma_{\text{LLM}}^{\text{logic}}(a_i) \right),
  \label{eq:important}
\end{equation}
where $S_i$ is the overall score for the $i$-th answer, and $\Gamma_{\text{LLM}}^{\text{type}}(a_i)$, $\Gamma_{\text{LLM}}^{\text{loc}}(a_i)$, and $\Gamma_{\text{LLM}}^{\text{logic}}(a_i)$ represent the individual evaluation scores for defect type recognition, localization precision, and reasoning quality, respectively. These scores provide fine-grained supervision signals for ranking and optimization.

\begin{figure}[t]
  \centering
  \includegraphics[width=\linewidth]{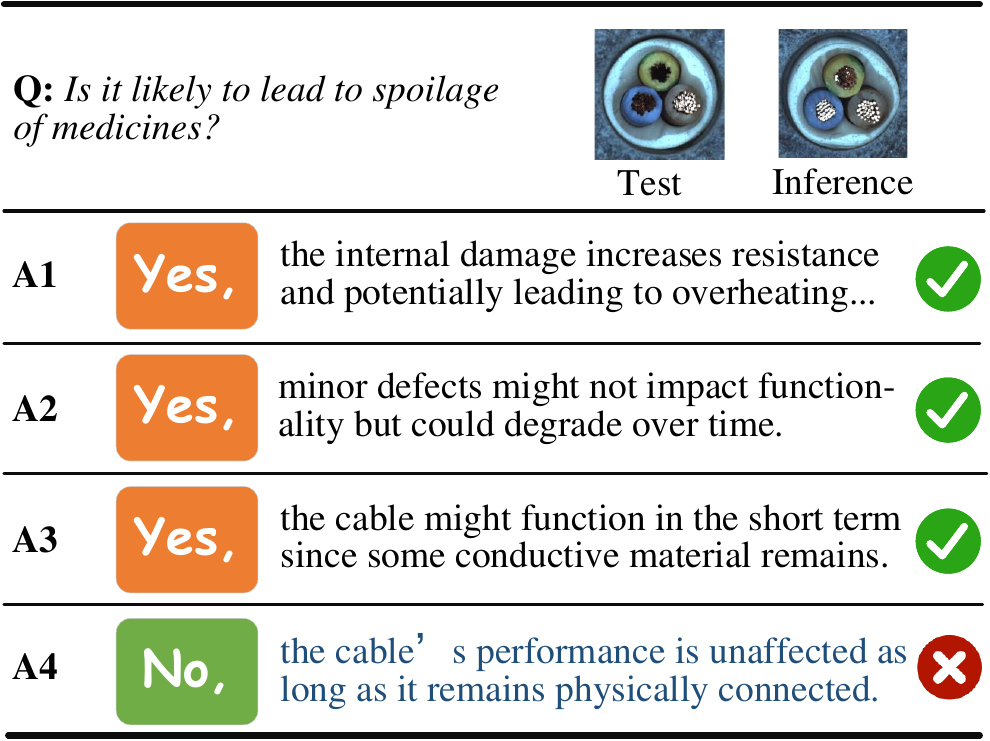}
  \caption{An example from the AD-PL dataset. Each instance consists of a test-reference image pair and a corresponding question, accompanied by four preference-ranked answers (A1 to A4), with the responses becoming progressively less appropriate or increasingly erroneous from A1 to A4.}
  \label{fig:dpo}
  \Description{A woman and a girl in white dresses sit in an open car.}
\end{figure}

We compute the entropy of these candidate answers to quantify the uncertainty and diversity among them. To this end, we normalize the MLE scores using a softmax function and calculate the entropy, defined as:
\begin{equation}
  H_i = - \sum_{i=1}^{L_N} \text{softmax}(S_i) \log \text{softmax}(S_i),
  \label{eq:important}
\end{equation}
where $H_i$ denotes the entropy of the $i$-th answer. This entropy reflects the confidence and informativeness of the model’s predictions. Lower entropy indicates more confident predictions, while higher entropy suggests ambiguity or conflicting signals.

Next, the Answer Selection step computes the most informative answer pairs based on entropy differences. Specifically, it selects the pair of adjacent ranked answers with the smallest entropy difference:
\begin{equation}
  \hat{i} = \arg \min_{1 \leq i \leq L_N-1} \left| H(a_i) - H(a_{i+1}) \right|,
  \label{eq:important}
\end{equation}
where $\hat{i}$ denotes the index of the answer pair $(a_i, a_{i+1})$ that exhibits the minimal entropy difference, denoted as $\Delta H_{min}$. Similarly, the answer pair with the maximal entropy difference, denoted as $\Delta H_{max}$, corresponds to $(a_1, a_{L_N})$. A large entropy difference reflects significant variation in answer quality, while a small entropy difference captures subtle distinctions, serving as fine-grained alignment.
To ensure the model learns both the differences and similarities between ranked answers, we incorporate a bias term into the Bradley–Terry framework \cite{bradley1952rank}. The modified objective is defined as:
\begin{equation}
  d(x, a_w, a_l, \Delta H) = \log \pi_\theta(a_{w} | x) - \log \pi_\theta(a_{l} | x) - \eta \Delta H,
  \label{eq:important}
\end{equation}
where $\Delta H \in \{\Delta H_{max}, \Delta H_{min}\}$ denotes the entropy difference between the winning answers $a_w$ and the losing answers $a_l$, and $\eta$ is a scaling factor.

Finally, in the Optimization stage, E-DPO fine-tunes the policy model $\pi_\theta$ (including the Adapter and LLM) against the frozen model $\pi_{\text{ref}}$ (obtained after SFE fine-tuning) using an entropy-aware loss $\mathcal{L}_{\text{E-DPO}}$, defined as:
\begin{align}
  &\mathcal{L}_{EDPO}(\pi_\theta) = - \mathbb{E}_{T}  \left[ \log \left( \frac{1}{1 +  \exp(-d(T))} \right) \right],
  \label{eq:dpo}
\end{align}
where $T=\{x, a_w, a_l, \Delta{H}\}$, and the $\Delta H_{\text{max}}$ encourages learning from distinguishable preference answers, while $\Delta H_{\text{min}}$ refines the decision of ambiguous examples. $\pi_{\text{ref}}$ is used to stabilize training via KL regularization. This formulation jointly enhances preference learning and alignment with expert preferences.

\subsection{Training and Inference}

\textbf{Training.} We adopt a two-stage training procedure. In the first stage (SFE), we train the adapter, the fusion attention layer, and the LLM to equip the model with domain knowledge of anomalies and comparison-based reasoning capabilities. In the second stage (E-DPO), the fusion attention layer is frozen, and further fine-tuning is applied to the adapter and LLM. The training objective for the LLM employs the standard cross-entropy loss, defined as:
\begin{equation}
  \mathcal{L}_{\text{ce}} = - \sum_{i=1}^{n} a_i \log(p_i),
  \label{eq:important}
\end{equation}
where $n$ is the number of tokens, $a_i$ is the true label for token $i$, and $p_i$ is the predicted probability. For E-DPO, we employ the maximum likelihood estimation as the training objective for a policy model parameterized by $\pi_{\theta}$, as defined in Eq.~\ref{eq:dpo}.

\noindent\textbf{Inference.} During the inference phase, the comparison-enhanced embedding obtained from the fusion attention layer is projected into a unified feature space via layer-normalized linear projections, formulated as:
\begin{equation}
    \mathbf{t}^{\text{input}} = \left[ \phi_{Proj}(\mathbf{v}); \phi_{Proj}(\mathbf{t}^s); \phi_{Proj}(\mathbf{t}^q) \right]
\end{equation}
where $\phi_{Proj}(\cdot)$ denotes a LayerNorm projection head. This unified representation is then fed into the LLM for auto-regressive generation as follows:
\begin{equation}
  p(Y_a | \mathcal{I}, Y_q) = \prod_{t=1}^{L} p_\theta(y_t | \mathcal{I}, Y_{q, <t}, Y_{a, <t}),
  \label{eq:important}
\end{equation}
where $Y_{a,<t}$ and $Y_{q,<t}$ denote the answer and question tokens from all previous turns. Finally, the model decodes the generated token sequence $Y_a = [y_1, y_2, \dots, y_L]$ into a coherent natural language response.

\begin{table*}[t]
\centering
\caption{Comparison of different methods on the MANTA and MPDD anomaly reasoning QA datasets under zero-shot and one-shot settings. Metrics include SBERT and accuracy across difficulty levels.}
\setlength{\tabcolsep}{3pt}
\begin{tabular}{l c c c c c c c c c c}
\toprule
 & \multicolumn{5}{c}{\textbf{MANTA \cite{fan2024manta}}} & \multicolumn{5}{c}{\textbf{MPDD QA \cite{9631567}}}\\
\cmidrule(lr){2-6} \cmidrule(lr){7-11}
 Method & Standard & Identification & Reasoning & SBERT & Average & Standard & Identification & Reasoning & SBERT & Average\\
\midrule
 LLaVA-1.6 \cite{liu2024llavanext}     & 65.3/45.3 & 53.4/43.2 & 63.8/64.2 & 0.59/0.55 & 59.0/49.0 & 72.3/60.9 & 39.7/49.3 & 61.7/56.4 & 0.64/0.67 & 57.1/55.3 \\
 Molmo \cite{molmo2024}         & 60.8/63.6 & 49.7/57.0 & 56.5/57.8 & 0.64/0.67 & 54.2/58.8 & 62.6/68.4 & 44.8/55.3 & 58.5/63.8 & 0.59/0.62 & 54.7/62.3 \\
 Qwen2-VL \cite{Qwen2-VL}      & 60.0/58.8 & 50.7/50.0 & 60.1/58.3 & 0.58/0.54 & 55.3/54.3 & 74.3/66.3 & 51.0/53.6 & 49.5/61.9 & 0.71/0.65 & 59.8/60.3 \\
 DeepSeek-VL2 \cite{wu2024deepseekvl2mixtureofexpertsvisionlanguagemodels}  & 56.2/58.9 & 46.0/49.2 & 42.0/44.2 & 0.53/0.63 & 47.5/50.4 & 57.2/60.4 & 46.9/48.8 & 57.1/56.3 & 0.60/0.62 & 53.1/55.0 \\
 GPT-4o \cite{hurst2024gpt}        & 58.4/60.1 & 57.4/59.3 & 75.4/76.1 & \textbf{0.71}/0.73 & 62.1/63.7 & 75.5/71.8 & \textbf{64.7}/58.1 & 62.5/67.0 & 0.74/0.78 & 68.5/65.4 \\ 
 InternVL2 \cite{chen2024internvl}       & 60.6/58.3 & 45.8/51.7 & 54.0/51.1 & 0.62/0.61 & 51.5/53.2 & 54.3/60.3 & 49.3/48.7 & 59.2/56.2 & 0.67/0.71 & 53.4/54.9 \\
 InternVL2 (sft) & 62.5/82.6 & 62.4/\textbf{65.7} & 70.1/71.4 & 0.67/0.68 & 64.3/71.3 & 70.6/78.9 & 60.7/58.7 & 63.0/65.1 & 0.73/0.71 & 65.1/68.3 \\
 SAGE (Ours)   & \textbf{70.6}/\textbf{84.6} & \textbf{65.7}/64.9 & \textbf{76.6}/\textbf{79.6} & 0.70/\textbf{0.73} & \textbf{69.6}/\textbf{73.5} & \textbf{81.6}/\textbf{83.3} & 61.6/\textbf{64.3} & \textbf{67.6}/\textbf{69.2} & \textbf{0.78}/\textbf{0.81} & \textbf{73.8}/\textbf{74.1} \\ 
\bottomrule
\end{tabular}
\label{tab:combined-shot}
\end{table*}

\begin{figure}[t]
  \centering
  \includegraphics[width=1\linewidth]{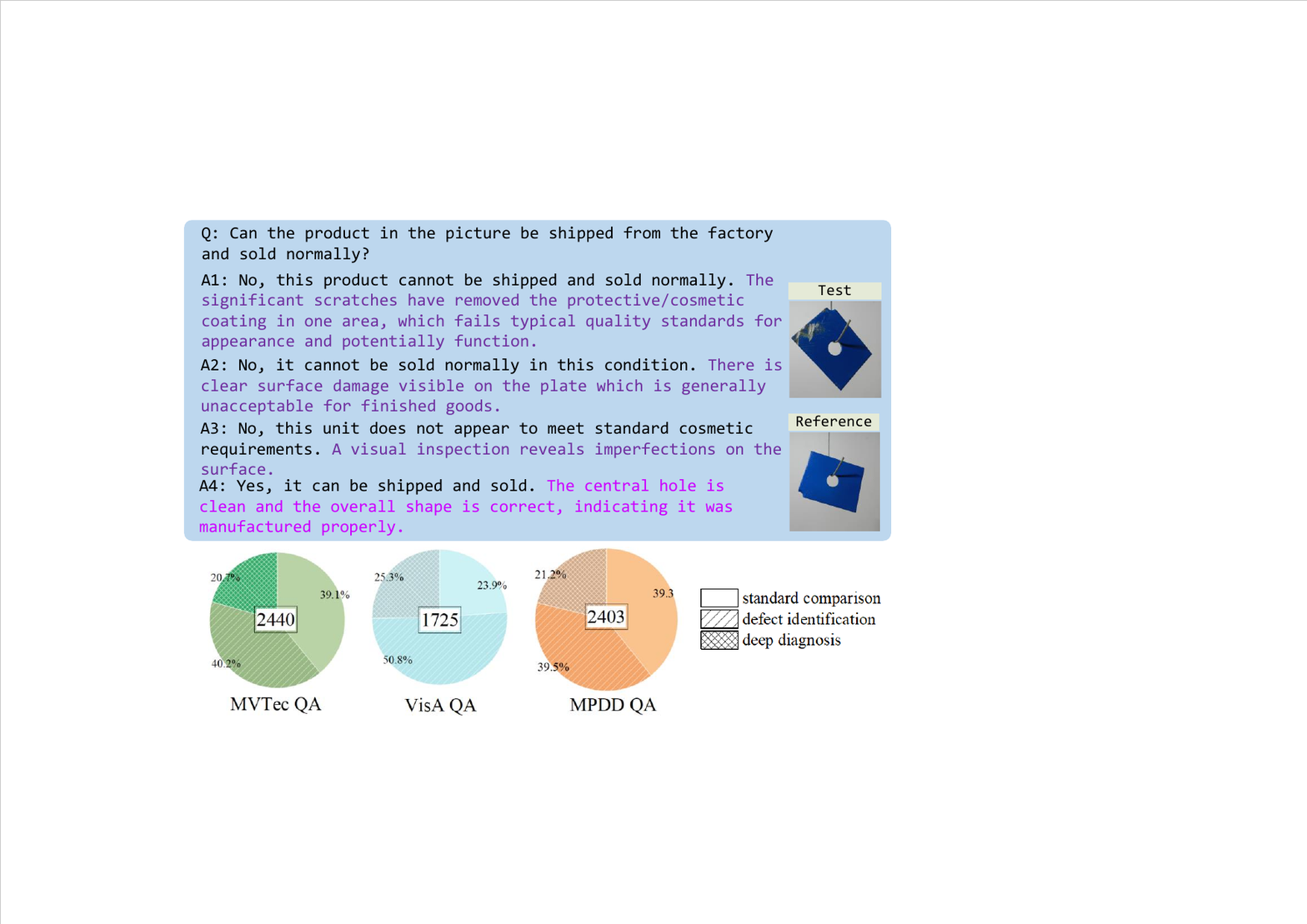}
  \caption{Example and overview of AD-PL dataset.}
  \label{label:anomaly_pref}
  \Description{For chapter 4}
\end{figure}

\section{AD-PL}
\label{sec:dataset}

To support preference learning and enhance the logical consistency and robustness of anomaly reasoning, we construct a new dataset, \textbf{AD-PL}, based on existing benchmarks: MVTec AD~\cite{bergmann2019mvtec}, VisA~\cite{zou2022spot}, and MPDD~\cite{9631567}. We adopt a three-stage pipeline for data construction. First, GPT-4o is employed to generate object-specific image captions, which are manually verified for accuracy and relevance. Second, we manually design a diverse set of anomaly scenarios and question templates, covering a range of anomaly types and reasoning complexities. Finally, these templates and contextual cues are input into GPT-4o to generate structured question-answer pairs, each containing four preference-ranked responses.

The AD-PL dataset comprises 28,415 QA instances, each featuring four preference-ranked answers. Test images are selected from MVTec QA (1,722), VisA QA (2,403), and MPDD QA (448), with a total of 1,788 normal and 2,775 anomaly images. The questions are categorized into three reasoning difficulty levels: 12,460 standard comparison instances, 10,635 defect identification cases, and 5,320 deep diagnostic cases. The dataset information and examples of AD-PL are illustrated in Figure~\ref{label:anomaly_pref}.

\section{Experiments}
\subsection{Setup}
We evaluated our method on two benchmarks: the public MANTA dataset and our proprietary MPDD QA dataset. MANTA is a dataset designed for multi-view anomaly detection, which categorizes questions into three difficulty levels: standard comparison, defect identification, and deep diagnosis. Each QA instance includes a test-reference image pair. MPDD QA is an anomaly reasoning dataset derived from MPDD, featuring 2,240 QA instances with test–reference image pairs categorized into the same three levels as MANTA.

We utilized the MVTec QA dataset for stage 1 training and the VisA QA dataset for stage 2 training. The backbone of SAGE is InternVL2, a powerful open-source VLM known for its strong multimodal capabilities. The vision encoder is InternViT‑6B‑448px‑V1‑5, while the language model is Nous‑Hermes‑2‑Yi‑34B. The training employed the AdamW optimizer with a peak learning rate of 2e-6 and a batch size of 32, distributed across 8 A100 GPUs. To stabilize the initial phase of training, we utilized a linear learning rate warmup for the first 1000 steps. Additionally, gradient clipping with a maximum norm of 1.0 was applied to prevent exploding gradients. To accelerate training and reduce memory consumption, mixed precision was enabled using the bfloat16 format.

\begin{table}[t]
\centering
\caption{Comparison of model performance on the MANTA dataset under zero-shot and one-shot settings across defect type, localization, and reasoning.}
\resizebox{\linewidth}{!}{
\setlength{\tabcolsep}{2pt}
\begin{tabular}{lcccc} 
\toprule
Method         & Defect Category & Defect Location & Reasoning Logic \\ 
\midrule
LLaVA-1.6 \cite{liu2024llavanext}     & 3.12 / 3.25     & 2.98 / 3.10     & 3.75 / 3.98     \\
Molmo \cite{molmo2024}         & 3.25 / 3.42     & 3.15 / 3.28     & 4.08 / 4.20     \\
DeepSeek-VL2 \cite{wu2024deepseekvl2mixtureofexpertsvisionlanguagemodels}  & 3.08 / 3.51     & 3.22 / 3.05     & 3.95 / 4.24     \\
Qwen2-VL \cite{Qwen2-VL}      & 3.30 / 3.12     & 3.18 / 3.31     & 3.85 / 4.13     \\
InternVL2  \cite{chen2024internvl}     & 3.45 / 3.47     & 3.30 / 3.22     & 3.70 / 3.41     \\
SAGE (Ours)          & \textbf{4.05 / 4.09} & \textbf{3.75 / 3.68} & \textbf{4.35 / 4.47} \\
\bottomrule
\end{tabular}
}
\label{tab:performance_comparison_transposed} 
\end{table}

We evaluated eight state-of-the-art baselines under both zero-shot and one-shot settings: LLaVA-1.6 \cite{liu2024llavanext} (7B), Molmo \cite{molmo2024} (7B), Qwen2-VL \cite{Qwen2-VL} (7B), DeepSeek-VL2 \cite{wu2024deepseekvl2mixtureofexpertsvisionlanguagemodels} (4.5B), GPT-4o \cite{hurst2024gpt}, InternVL2 \cite{chen2024internvl} (7B), and its fine-tuned variant. In the zero-shot setting, the encoded test image features are reused as both keys and values in the fusion attention mechanism during inference. Following the MANTA benchmark, we adopted accuracy as the primary metric. To further validate robustness, we introduced Accuracy-for-Group (Acc-G), a metric that requires correct answers to all five interrelated questions. For the open-ended responses, while conventional approaches relied on methods such as SBERT \cite{reimers2019sentence} to measure semantic similarity, we argued that such methods fail to capture task-specific reasoning quality \cite{zheng2023judging}. To address this, we introduced a framework to evaluate model capabilities across three aspects. For this evaluation, we developed an evaluation pipeline using GPT-4o, which assessed model-generated responses and assigned a quantitative score on a scale of 1-5 based on the following criteria:
\begin{itemize}[topsep=0pt, parsep=0pt, itemsep=1pt]
    \item Anomaly Identification: Distinguishing anomaly regions.  
    \item Anomaly Localization: Identifying boundaries of anomalies.
    \item  Logical Consistency in Reasoning: Logical coherence and factual correctness of explanations.
\end{itemize}

\begin{figure}[t]
  \centering
  \includegraphics[width=\linewidth]{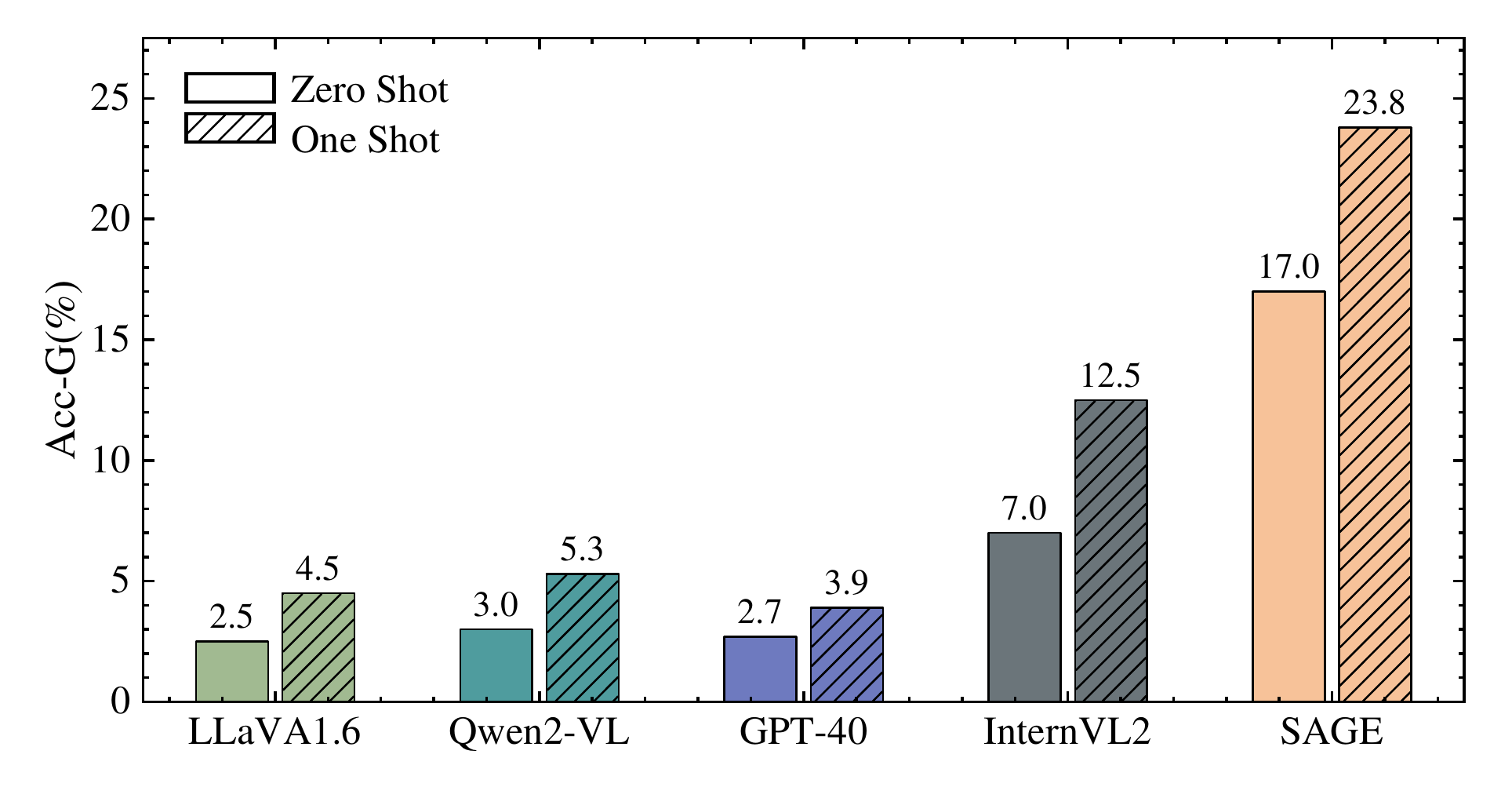}
  \caption{Accuracy results by group on the MANTA dataset in the one-shot setting.}
  \label{label:group result}
\end{figure}

\begin{figure*}[t]
    \centering
    \subfigure[Result for GPT-4o and SAGE in a zero-shot setting]{
        \includegraphics[width=0.49\textwidth]{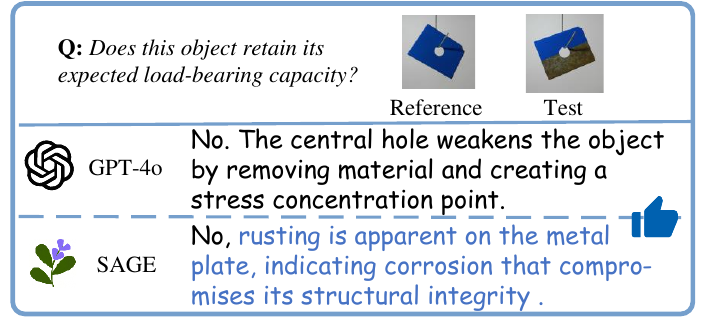}
        \label{fig:a}
    }
    \subfigure[Result for GPT-4o and SAGE in a one-shot setting]{
        \includegraphics[width=0.48\textwidth]{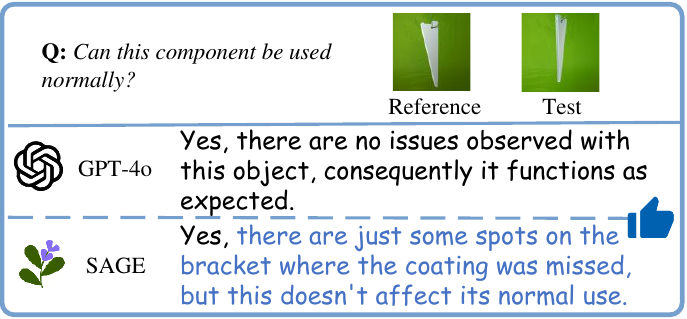}
        \label{fig:b}
    }
    \\
    \subfigure[Result for GPT-4o and SAGE in a zero-shot setting]{
        \includegraphics[width=0.48\textwidth]{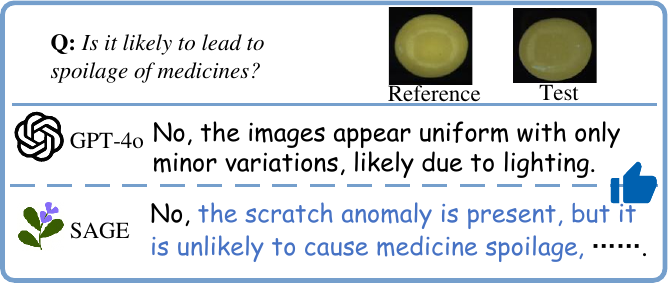}
        \label{fig:c}
    }
    \subfigure[Result for GPT-4o and SAGE in a one-shot setting]{
        \includegraphics[width=0.49\textwidth]{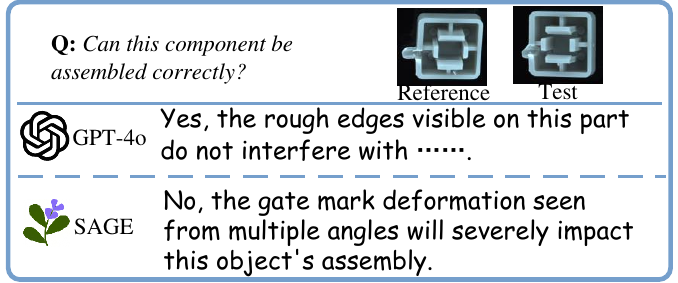}
        \label{fig:d}
    }
    \caption{Qualitative comparison between GPT-4o and SAGE. The appropriate and inappropriate parts of the responses are highlighted. SAGE correctly identified the anomaly and provided comprehensive reasoning.}
    \label{fig:qualitative result}
\end{figure*}

\subsection{Quantitative Evaluation}
\textbf{Accuracy Evaluation.}
We conducted experiments to evaluate the performance of our SAGE against several state-of-the-art baseline models on the anomaly reasoning task. The evaluation was performed on the MANTA and MPDD QA benchmark datasets under zero-shot and one-shot settings. Performance was measured using Accuracy and SBERT scores for responses, as well as the Acc-G metric to assess group-level reasoning capabilities in the one-shot setting. Detailed comparative results are presented in Table~\ref{tab:combined-shot}, showing that SAGE has outstanding performance in different level sub-tasks compared to baselines. 

On the MANTA dataset under zero-shot setting, SAGE achieves an Average Accuracy of 69.6\%, surpassing GPT-4o baseline by 7.5\%. In the corresponding one-shot setting on MANTA, SAGE reaches 73.5\% Average Accuracy, exceeding InternVL2 (sft), by 2.2\%. A similar strong performance is observed on the MPDD QA dataset. Complementing these findings, the group-level evaluation presented in Figure~\ref{label:group result} also highlights SAGE's substantial advantage; it achieves approximately 17\% Acc-G in the zero-shot setting and nearly 24\% in the one-shot setting, further validating SAGE's superior capability in this aspect. This strong anomaly reasoning capability stems from learning generalizable normal/abnormal semantics via the AD-PL dataset, coupled with the enhanced generalization provided by the Fact Enhancement. Together, these elements enable accurate and comprehensive anomaly analysis.

\noindent\textbf{Generation Performance Evaluation.}
To evaluate the quality of generated responses in anomaly analysis, we conducted a generation assessment using the MLE framework. The evaluation focused on three key aspects: accuracy in identifying the Defect Category, precision in locating the Defect Location, and coherence of the Reasoning Logic. We compared our model, SAGE, with baselines under both zero-shot and one-shot settings. Results are summarized in Table~\ref{tab:performance_comparison_transposed}. SAGE consistently outperformed all baselines across all criteria and settings. Notably, it achieved the highest scores in Reasoning Logic with 4.35 (zero-shot) and 4.47 (one-shot), and led in Defect Category identification with 4.05 and 4.09. These results highlight SAGE's capability in generating informative, high-quality responses. Its strong performance in Reasoning Logic indicates well-structured explanations, while its accuracy in Defect Category and Location reflects a robust understanding of anomaly details.

\subsection{Ablation Results}
To validate the contributions of our proposed method, we conducted an ablation study in the zero-shot setting on MANTA, starting from an InternVL2 baseline (results in Table~\ref{tab:ablation}). We first evaluated the SFE. Compared to simply concatenating facts (`+Fact[Hard]'), which served as a basic comparison method, our FE (`+Fact[FE]') yielded significantly better results, improving Accuracy by 4.86\% and Acc-G by 2.32 points, confirming the efficacy of its interactive mechanism. Subsequently, we assessed the impact of our E-DPO technique by comparing the full model against a version using only standard DPO ('FE+DPO'). The full model achieved substantial gains, notably boosting Acc-G by 4.4 points and the overall Score by 0.26 points. This clearly demonstrates the significant added value of E-DPO in enhancing both overall performance and response quality beyond standard preference alignment. Thus, the ablation study confirms the effectiveness of both the SFE and E-DPO.

\begin{table}[t]
\centering
\caption{Ablation results on MANTA under the zero-shot setting. Using InternVL2 as the baseline, we compare direct fact concatenation (Fact[Hard]) with fusion attention enhancement (Fact[FE]). Based on Fact[FE], we assess DPO (FE+DPO) and E-DPO (Full model) for preference optimization.}
\begin{tabular}{lccccc}
\hline
Method & Accuracy & SBERT & Acc-G & Score\\
\hline
Baseline & 51.51 & 0.62 & 1.25 & 3.48\\
+Fact[Hard]  & 57.37 & 0.61 & 7.36 & 3.44\\
+Fact[FE] & 62.23 & 0.64 & 9.68 & 3.65\\
+DPO & 60.05 & 0.67 & 7.29 & 3.62\\
SFE+DPO & 67.28 & 0.67 & 12.92 & 3.79\\
Full Model & 69.64 & 0.70 & 17.32 & 4.05\\
\hline
\end{tabular}
\label{tab:ablation}
\end{table}

\begin{figure}[t]
    \centering
    \subfigure[Accuracy results.]{%
        \includegraphics[width=0.42\linewidth]{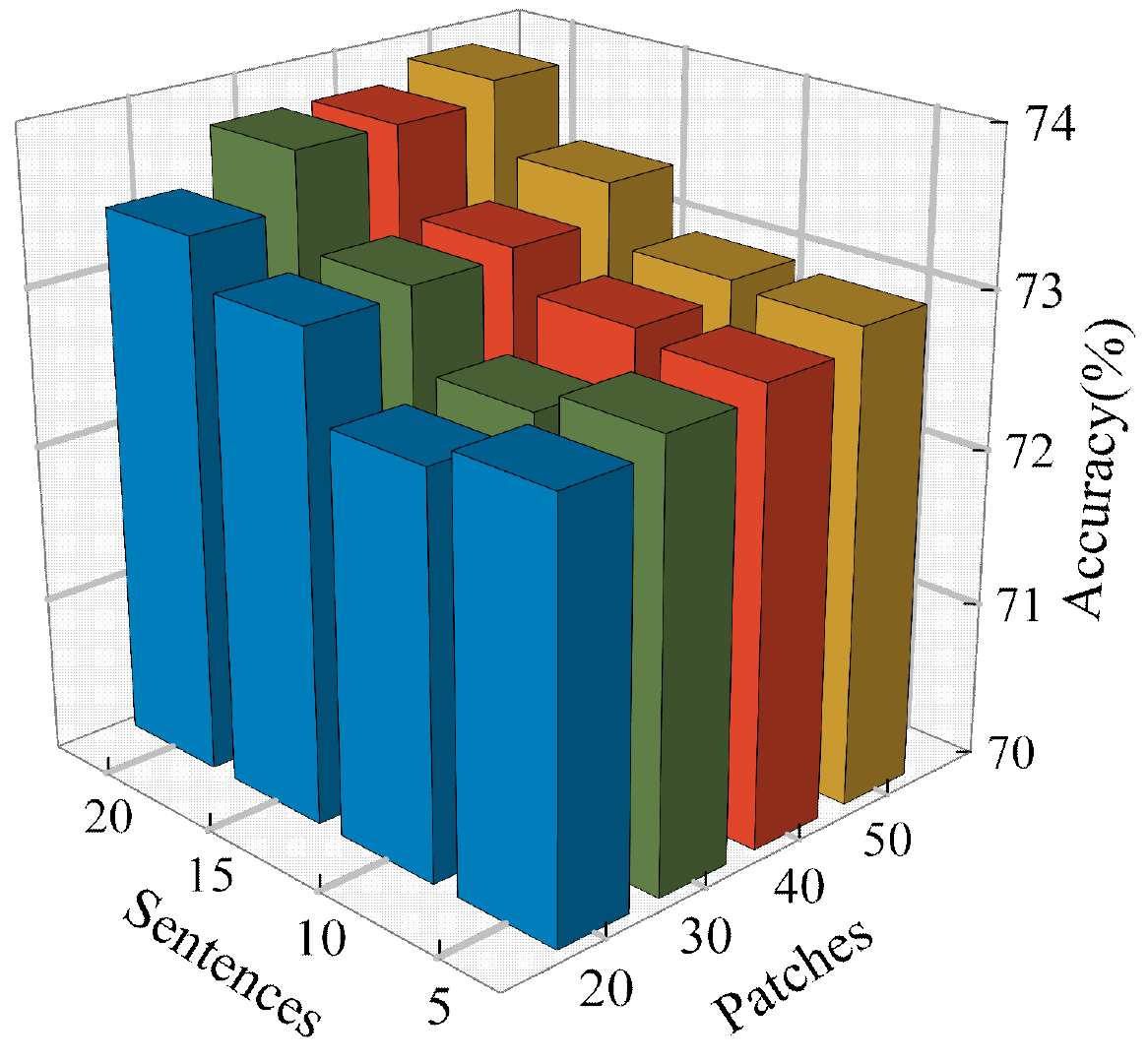}
    }\hspace{0.2cm} 
    \subfigure[Score results.]{%
        \includegraphics[width=0.42\linewidth]{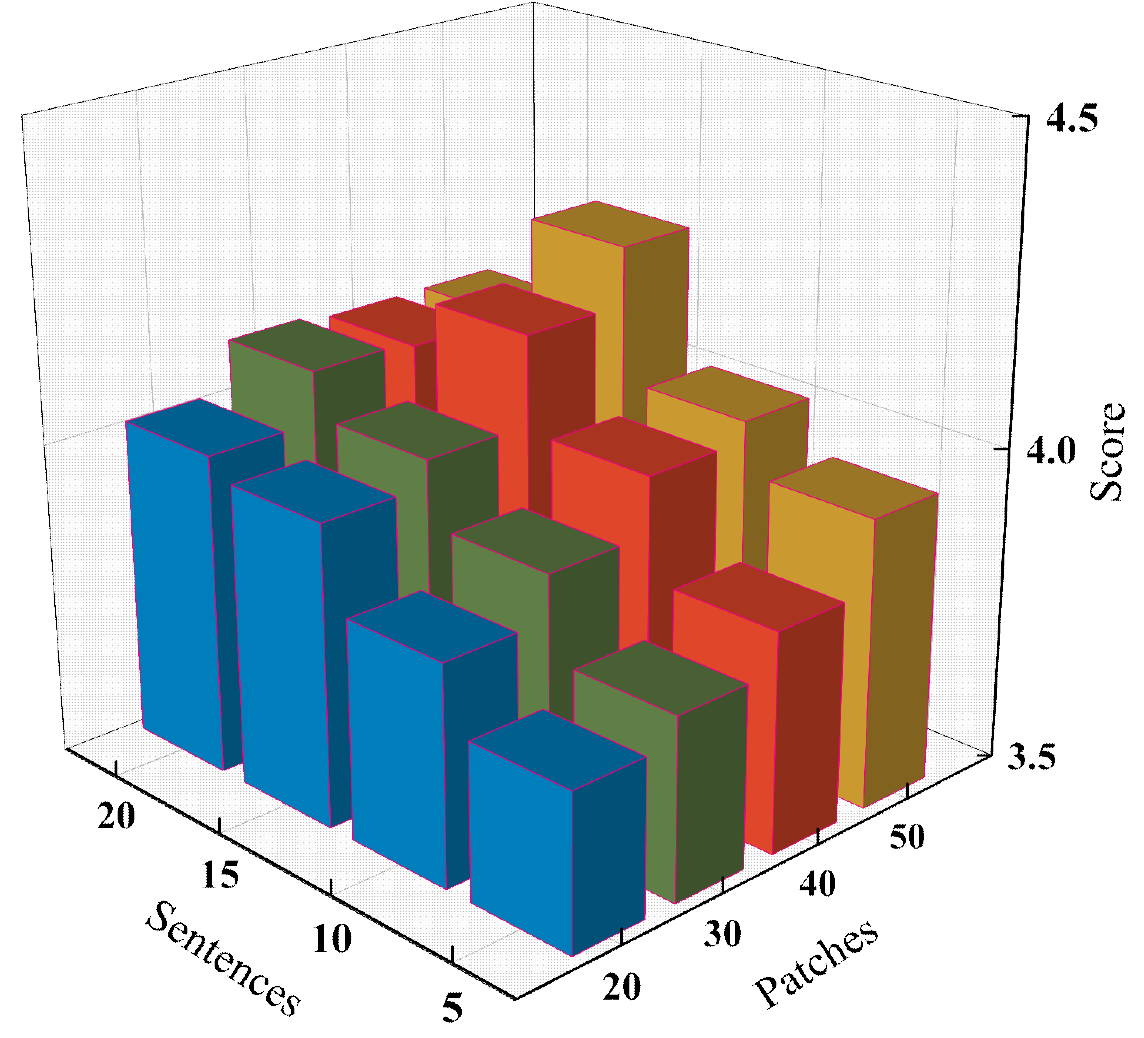}
    }
    \caption{Accuracy and score results on MANTA in the 1-shot setting using fact sentences and distinct patches.}
    \label{fig:double_images}
\end{figure}

We analyzed the effect of sentences and patches in the 1-shot setting, which represents the number of fact sentences, along with the distinct top-k patches between test and reference images in the SFE. Figure~\ref{fig:double_images} illustrates the effect of varying the number of fact sentences and distinct top-k patches on model performance. In terms of accuracy, the impact of Patches is observed to be relatively minor, with negligible differences in accuracy between N=20 and N=50. However, sentences exhibit a more pronounced influence, with accuracy demonstrating a clear positive correlation with an increasing number of sentences. In terms of the Score metric, an increase in both Patches and Sentences generally correlates with an improvement in the model's average score.



\subsection{Qualitative Examples}
Our model demonstrates the ability to generate deep reasoning. Figures~\ref{fig:qualitative result} illustrate the performance of SAGE on anomaly reasoning tasks evaluated under zero-shot and one-shot conditions. Notably, the quality of the generated responses is enhanced in the one-shot setting compared to the zero-shot setting, attributable to the guidance provided by the reference object in the one-shot setup.

\section{Conclusion}
In this paper, we establish a large-scale anomaly reasoning benchmark, AD-PL, to address the data scarcity issue for visual anomaly reasoning. We then introduce SAGE, a specialist VLM featuring two key components: self-guided fact enhancement, which guides the LLM's visual focus to improve detailed comparison capabilities, and E-DPO, which aligns the response with human preference. Consequently, SAGE achieves significant improvements in overall anomaly reasoning compared to prior approaches.

\section*{Limitations} While SAGE demonstrates strong performance, its effectiveness relies heavily on the quality of the SFE module. As this module generates textual facts without external supervision, potential inaccuracies or biases may impair reasoning or misguide visual attention. Moreover, although the AD-PL dataset offers a diverse benchmark for anomaly reasoning, it may not fully capture real-world complexities such as domain shifts, rare anomalies, or multi-modal inconsistencies, which can hinder generalization.

\section*{Acknowledgments}
This work was supported by the Key Research and Development Program of Heilongjiang Province of China (Grant No. 2022ZX01A21), and the State Key Laboratory of Massive Personalized Customization System and Technology, COSMOPlat Institute of Industrial Intelligence (Qingdao) Co. Ltd., located in Qingdao 266100, China.
\bibliographystyle{ACM-Reference-Format}
\bibliography{reference}

\end{document}